\documentclass[mlabstract]{jmlr}

\usepackage{longtable}

\usepackage{booktabs}
\usepackage[load-configurations=version-1]{siunitx} 

\usepackage{wrapfig}
\usepackage{bm}
\usepackage{enumitem}
\setlist{
    topsep=0pt,
    partopsep=0pt,
    nosep
}
\usepackage[nameinlink,capitalise,noabbrev]{cleveref}

\theorembodyfont{\upshape}
\theoremheaderfont{\scshape}
\theorempostheader{:}
\theoremsep{\newline}

\jmlrvolume{}
\firstpageno{1}
\editors{\footnotesize{Sophia Sanborn, Christian Shewmake, Simone Azeglio, Arianna Di Bernardo, Nina Miolane}}

\jmlryear{2022}
\jmlrworkshop{NeurIPS Workshop on Symmetry and Geometry in Neural Representations}

\title{Learning unfolded networks with a cyclic group structure}

\author{\Name{Emmanouil Theodosis} \Email{etheodosis@g.harvard.edu}\\
\addr School of Engineering and Applied Sciences\\
Harvard University\\
Cambridge, MA 02138
\AND
\Name{Demba Ba} \Email{demba@seas.harvard.edu}\\
\addr School of Engineering and Applied Sciences\\
Harvard University\\
Cambridge, MA 02138}

\begin{document}

\maketitle

\begin{abstract}
Deep neural networks lack straightforward ways to incorporate domain knowledge and are notoriously considered black boxes. Prior works attempted to inject domain knowledge into architectures \emph{implicitly} through data augmentation. Building on recent advances on equivariant neural networks, we propose networks that \emph{explicitly} encode domain knowledge, specifically equivariance with respect to rotations. By using unfolded architectures, a rich framework that originated from sparse coding and has theoretical guarantees, we present interpretable networks with sparse activations. The equivariant unfolded networks compete favorably with baselines, with only a fraction of their parameters, as showcased on (rotated) MNIST and CIFAR-10.
\end{abstract}
\begin{keywords}
Equivariance, model-based learning, cyclic groups, unfolded networks
\end{keywords}

\section{Introduction}
\label{sec:intro}
While advances in deep neural networks have yielded groundbreaking results in various fields such as computer vision \citep{ReFa17, PZDD17, MST+20}, natural language processing \citep{DCL+19, BMR+20}, and their intersection \citep{RKH+21}, interpreting their structure and explaining their performance is not straightforward. At the same time, applying deep learning techniques to novel fields comes with challenges, as it is not clear how to integrate domain knowledge into existing architectures. In this work, we propose a novel architecture to address both of these shortcomings at the same time, leading to an interpretable architecture that respects expert knowledge.

\begin{figure}[b!]
    \centering
    \includegraphics[scale=0.5]{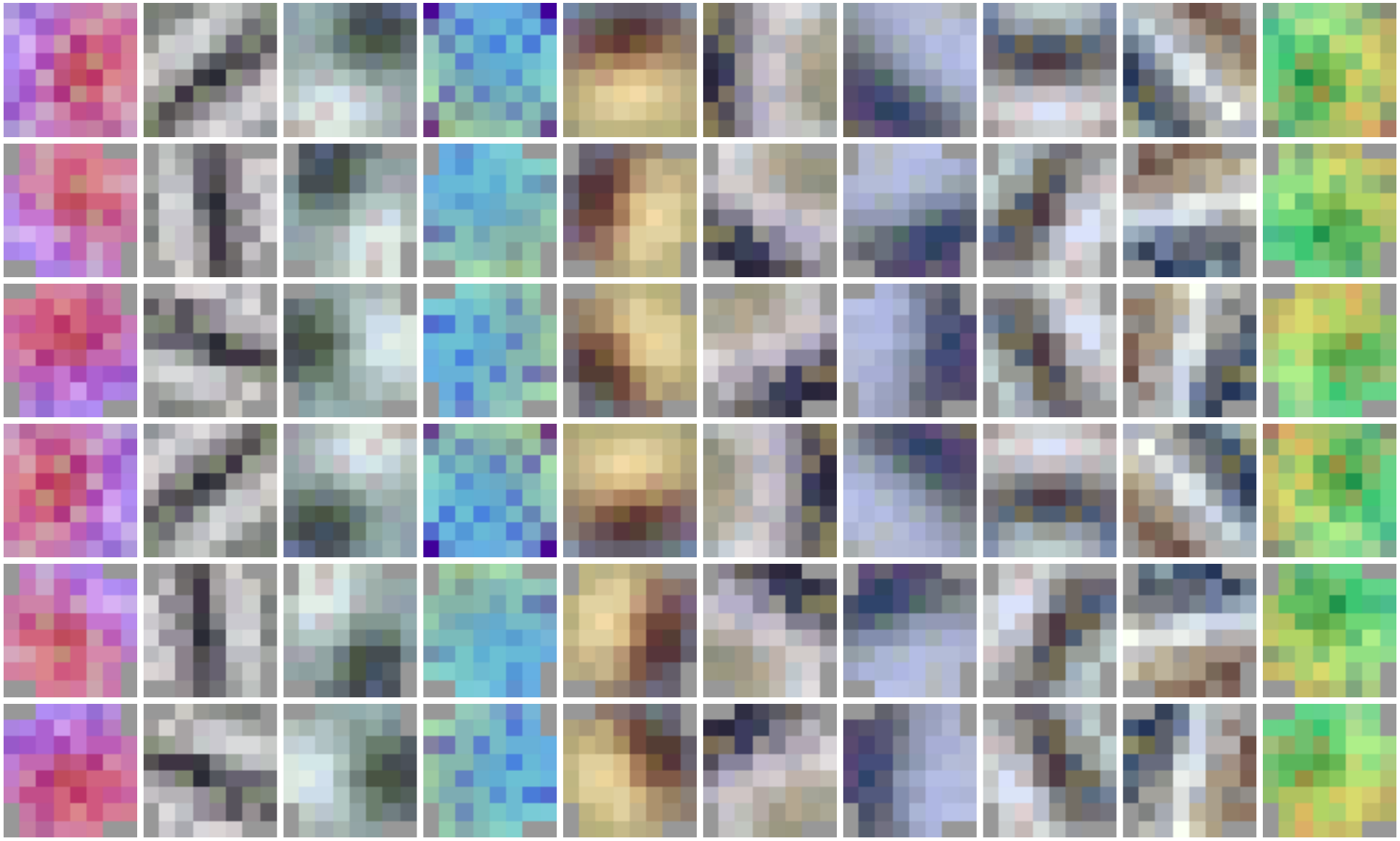}
    \caption{Filters learned at the final layer of $R_{60}$-CNN, training on CIFAR-10.}
    \label{fig:intro}
\end{figure}

Convolutional Neural Networks (CNNs) are \emph{equivariant} in their representations with respect to translation; however, there are other operators that is natural for image models to be equivariant to, such as rotations. While data augmentation techniques have been used to model equivariances they require large amounts of data and increase the computational demands for training, frequently by an order of magnitude. At the same time, if we know the desired equivariances for a specific application, investing computational resources to re-learn these equivariances is wasteful. This was also acknowledged by \citet{DDK16} and \citet{CoWe16} who concurently introduced CNN frameworks that incorporate rotated filters in order to create equivariant representations with respect to rotations; however both works were limited to elementary rotations and the reflections of $\mathcal{D}_4$. Follow up works by different authors extended the ideas to vector fields \citep{MVKT17}, applied rotations directly on the sphere to avoid interpolation artifacts \citep{EMD20}, and incorporated harmonic functions to model equivariance with respect to arbitrary rotations \citep{WGTB17}.

There have been several attempts to tackle interpretability, ranging from prototype learning approaches \citep{CLT+19, ArPf20} that learn \emph{prototypical} parts for each class in a classification task, to post-hoc methods \citep{RSG16} that reverse engineer predictions from arbitrary classifiers. In this work, we focus on \emph{model-based} networks \citep{SWED20}: in these approaches, interpretability is directly encoded into the model by constructing a neural network to mimic the steps of an optimization algorithm. First introduced by \citet{GrLe10}, \emph{unrolled} neural networks have inspired a vast array of works, ranging from theoretical contributions \citep{NWH19, AGM+15} to state-of-the-art results \citep{TST+20}.

In this work, we propose an unrolled architecture, inspired from algorithms for sparse coding, whose layer weights employ a \emph{cyclic group structure} to achieve rotational equivariance. This allows us to significantly reduce the number of trainable filters compared to baseline architectures. Concretely, our contributions can be summarized as follows:
\begin{enumerate}
    \item We propose an \emph{unrolled} architecture, modeled after sparse coding, that is by construction interpretable and equivariant with respect to rotations,
    \item we showcase its efficacy in learning filters that are governed by a cyclic group structure with a fraction of the learnable parameters, and
    \item we evaluate the proposed architecture on MNIST, rotated MNIST, and CIFAR-10, standard benchmarks for rotationally equivariant architectures and demonstrate its performance.
\end{enumerate}
We make the code for our experiments and architectures available online on GitHub at \url{https://github.com/manosth/cyclical_groups}.

\section{Background}
\label{sec:sec2}
\textbf{Equivariance.} In lay terms, an operator is equivariant with respect to some actions if it behaves in a predictable manner under them. Formally, we say that an operator $f$ is equivariant with respect to a family of actions $\mathcal{T}$ if, for any $T \in \mathcal{T}$ it holds that
\begin{equation}
    f(T(x)) = T'(f(x)),
\end{equation}
for some \emph{other} transform $T'$. Note that, in general, $T$ and $T'$ belong in different families as $T$ acts on the input space ($\operatorname{Dom}(f)$) and $T'$ on the encoding space ($\operatorname{Im}(f)$). Constant functions are trivially equivariant, and a special case, \emph{invariance}, arises when $T'$ is the identity map. Note that convolution is \emph{not} equivariant to rotation \citep{CoWe16, DDK16}; instead, the two are related by
\begin{equation*}
    R(\bm{x})*\bm{h} = R(\bm{x} * (R^{-1}(\bm{h})),
\end{equation*}
where $\bm{x}$ denotes an input image, $R$ is a rotation, and $\bm{h}$ is the convolving filter.
\\
\\
\textbf{Cyclic groups.} We call a finite group $G$ a cyclic group if there exists a generating element $g$ such that
\begin{equation}
    G = \{e, g, g^2, \ldots, g^{k-1}\},
\end{equation}
where $e$ denotes the identity element. We denote the family of cyclic groups as $\mathcal{G}$; several groups belong to this family, with most notable being $\mathcal{D}_4$, the symmetry group of the square. Cyclic groups are of interest for our model since all elements can be identified by the generator $g$. This will enable us, in \cref{sec:sec3} to significantly reduce the trainable parameters of our networks, while retaining (and even improving) performance and interpretability.
\\
\\
\textbf{Unrolled sparse autoencoders.} In their most general form, unrolled networks temporally unroll the steps of optimization algorithms, mapping algorithm iterations to network layers. In that way, the output of the neural network can be interpreted as the output of the optimization algorithm, with theoretical guarantees under certain assumptions. The \emph{Iterative Soft Thresholding Algorithn} (ISTA), an algorithm for sparse coding, has inspired several architectures \citep{SiEl19, SABE20, TDB21}, due to the desirability of sparse representations. Within that framework, the representation at layer $l + 1$ is given by
\begin{equation}
    \label{eq:ista}
    \bm{z}^{(l+1)} = \mathcal{S}_{\lambda}\left(\bm{z}^{(l)} + \alpha\bm{W}_l^T(\bm{x} - \bm{W}_l\bm{z}^{(l)})\right),
\end{equation}
where $\bm{x}$ is the \emph{original} input, $\bm{z}^{(l)}$ is the representation at the previous layer, $\bm{W}_{l}$ are the weights of layer $l$, $\alpha$ is a constant such that $\frac{1}{\alpha} \geq \sigma_{\max}(\bm{W}_l^T\bm{W}_l)$, and $\mathcal{S}_{\lambda}$ is the \emph{soft thresholding} operator defined as
\begin{equation}
    \mathcal{S}_{\lambda}(u) = \operatorname{sign}(u)\cdot\operatorname{ReLU}(\lvert u\rvert - \lambda).
\end{equation}
If the weights of all the $L$ layers are equal, i.e. $W_1 = \ldots = W_L$, we call the network \emph{tied}. As a final remark, \cref{eq:ista} can be rewriten as
\begin{equation}
    \bm{z}^{(l+1)} = \mathcal{S}_{\lambda}\left((I - \alpha\bm{W}_l^T\bm{W}_l)\bm{z}^{(l)} + \alpha\bm{W}_l^T\bm{x}\right) = \mathcal{S}_{\lambda}\left(\bm{W}_z \bm{z}^{(l)} + \bm{W}_x \bm{x}\right),
\end{equation}
where we let $\bm{W}_z = (I - \alpha\bm{W}_l^T\bm{W}_l)$ and $\bm{W}_x = \alpha\bm{W}_l^T$, which can be interpreted as a residual network \citep{HZRS16}, with a residual connection to the input.

\section{Equivariant autoencoders}
\label{sec:sec3}
We will combine the ideas of cyclic groups and unrolled autoencoders from \cref{sec:sec2} to create an \emph{equivariant} unrolled architecture, where the bulk of the weights in each layer are cyclic rotations of some ``basis'' weights . Let $R_{\theta}$ denote a two-dimensional rotation by $\theta$ degrees. If $360 \mod \theta = 0$ and we let $k = 360 \div \theta$, then the group defined as
\begin{equation}
    G = \{e, R_{\theta}, \ldots, R_{\theta}^{k-1}\},
\end{equation}
is a cyclic group generated by the generator $g = R_{\theta}$. This construction allows us to extend this framework, in future work, to \emph{arbitrary} linear operators and the possibility to \emph{learn} the generator $g$, leading to data-driven approaches for the cyclic group structure. Regardless, the weights of layer $l$ satisfy
\begin{equation}
    \label{eq:wl}
    \bm{W}_l = \begin{bmatrix}
                \bm{w}_l & R_{\theta}(\bm{w}_l) & \ldots & R_{\theta}^{k-1}(\bm{w}_l),\\
                \end{bmatrix}
\end{equation}
where $k$ is the number of rotations we are considering, per layer. Note that \cref{eq:wl} shows a single basis weight per layer; this can be readily extended to having $m$ basis weights $\bm{w}_{l_1}, \ldots, \bm{w}_{l_m}$ along with the $k$ rotations for each of them, as is done in \cref{sec:sec4}. The networks are constructed analogously to \citet{SABE20, TDB21} but with \emph{one} (or $m$) learnable filter(s) per layer. During training, the remaining filters are constructed and errors are backpropagated through \emph{all} filters. The experiments of \cref{sec:sec4} use untied networks for improved performance.

\section{Experiments}
\label{sec:sec4}
\begin{table}[b!]
	\centering
	\begin{tabular}{ccccc}
	\toprule
        Method & MNIST & rot-MNIST & CIFAR-10 & Number of parameters\\
        \toprule
		Baseline & \textbf{99.21} & 85.48 & 71.87 & 56.29K\\
        \midrule
        $R_{90}$-CNN & 99.12 & \textbf{86.62} & 72.20 & 21.73K\\
        $R_{60}$-CNN & 98.77 & 80.07 & \textbf{73.14} & \textbf{17.89K}\\
		\bottomrule
    \end{tabular}
    \caption{Performances of the baseline model, $R_{90}$-CNN, and $R_{60}$-CNN on different datasets along with their total number of trainable parameters (CIFAR-10).}
    \label{tab:results}
\end{table}
We present experiments on MNIST, rotated MNIST, and CIFAR-10, standard benchmarks for equivariant architectures, to evaluate the architecture introduced in \cref{sec:sec3}. We used batch normalization \citep{IoSz15} in all of our architectures, following best practices. The normalization was applied at the output of every layer, except the last. FISTA \citep{BT09}, a faster version of ISTA, is used for faster convergence of the sparse coding. All of our networks use $L = 4$ layers, $\lambda$ (the parameter of $\mathcal{S}_{\lambda}$) is set to $0.5$, and the stepsize of FISTA in \eqref{eq:ista} is set to $\alpha = 0.01$. A summary of our main results is given in \cref{tab:results}.

We test three models: an unrolled sparse network that is considered as a baseline; $R_{90}$-CNN, an equivariant unfolded network with the elementary rotations; and $R_{60}$-CNN, with $60\degree$ rotations. Every architecture has $60$ filters per layer; in the case of the baseline, all $60$ are learnable, $R_{90}$-CNN has $15$ learnable filters per layer the rest are constructed as elementary rotations, and $R_{60}$-CNN has only $10$ learnable filters per layer with the rest being their $60\degree$ rotations. The learned filters are $7\times 7$ in the case of MNIST and $8\times 8$ in the case of CIFAR-10. A visualization of $R_{60}$-CNN's learned filters when trained on CIFAR-10 is show in \cref{fig:intro}.

\textbf{MNIST.} We first evaluate all three models on MNIST, a relatively easy dataset, and find that all of them performed similarly. However, we note that $R_{90}$-CNN has only $\frac{1}{4}$ the trainable filters of the baseline, as the rest are generated as rotations of the basis weights; $R_{60}$-CNNH has only $\frac{1}{6}$. When evaluating the architectures on the rotated MNIST, a harder dataset, we observe that the $R_{90}$-CNN, with a \emph{fraction} of the parameters of the baseline model leads to the best performance. This showcases that the encoded equivariance in the representation is actually beneficial for the classification of the inputs.

Evaluating the architectures on the rotated MNIST produces similar results and $R_{90}$-CNN outperforms the baseline. However, $R_{90}$-CNN displays reduced performance on this dataset; we speculate that this is because of interpolation artifacts, which was one of the motivating reasons for \cite{EMD20} to consider rotations on the sphere.

To further demonstrate the benefit of the equivariant unfolded networks, we trained \emph{dense} variants (where the filters are now $28\times 28$ images) of the three models on MNIST, and
\begin{wraptable}{r}{6.5cm}
\vspace{-1em}
\begin{tabular}{ccc}\\
    \toprule
    Method & MNIST & rot-MNIST \\
    \toprule
    Baseline & 97.75 & 36.89\\
    \midrule
    $R_{90}$-NN & \textbf{98.04} & 37.02\\
    $R_{60}$-NN & 97.73 & \textbf{37.4}\\
    \bottomrule
\end{tabular}
\caption{Models trained on MNIST but evaluated on rot-MNIST.}
\label{wrap-tab:1}
\vspace{-1em}
\end{wraptable}
evaluate their performance on the rotated dataset. The results are presented in \cref{wrap-tab:1}. This experiment showcases the generalization capabilities of the equivariant unrolled networks. While we see similar performance across all models on the trained dataset, we see that \emph{both} the equivariant models are able to generalize better than the baseline. Dense architectures were chosen for this experiment to highlight the distribution shift when evaluating on rot-MNIST.

\textbf{CIFAR-10.} When training on CIFAR-10, a comparetively harder dataset, we found that \emph{both} the equivariant models outperform the baseline, even if marginally, with only a fraction of the learnable filters per layer. Moreover, upon further examination, the filters of $R_{90}$-CNN and $R_{60}$-CNN exhibit a topographic structure by construction as filters are generated as rotations of one another. That topographic structure is not present in the filters of the baseline model which doesn't incorporate domain knowledge regarding rotations. The learned filters of the different models, when trained on CIFAR-10, are presented in \cref{apd:first}.

\section{Conclusions and future work}
In this work we introduced equivariant unrolled networks, where the filters of each layer are constructed as discrete rotations of one another. By exploiting this cyclical group structure, we are able to facilitate training and maintain performance while drastically reducing the number of trainable parameters of the model. We experimentally validated the equivariant unrolled networks against a baseline network and showed comparable, and even favorable, results, with only a fraction of the learnable parameters on multiple datasets that are used as benchmarks for equivariant architectures. Finally, we identify \emph{learning} the generator $g$ from data, as hinted in \cref{sec:sec2}, an exciting avenue for future work.

\section*{Acknowledgements}
Demba Ba was supported by the National Science Foundation under Grant number \#S5206, PO 560825.

\bibliography{pmlr-sample}

\begin{thebibliography}{25}
\providecommand{\natexlab}[1]{#1}
\providecommand{\url}[1]{\texttt{#1}}
\expandafter\ifx\csname urlstyle\endcsname\relax
  \providecommand{\doi}[1]{doi: #1}\else
  \providecommand{\doi}{doi: \begingroup \urlstyle{rm}\Url}\fi

\bibitem[Arik and Pfister(2020)]{ArPf20}
Sercan Arik and Tomas Pfister.
\newblock Protoattend: Attention-based prototypical learning.
\newblock \emph{Journal of Machine Learning Research}, 21:\penalty0 1--35,
  2020.

\bibitem[Arora et~al.(2015)Arora, Ge, Ma, and Moitra]{AGM+15}
Sanjeev Arora, Rong Ge, Tengyu Ma, and Ankur Moitra.
\newblock Simple, efficient, and neural algorithms for sparse coding.
\newblock In \emph{Conference on Learning Theory}, 2015.

\bibitem[Beck and Teboulle(2009)]{BT09}
Amir Beck and Marc Teboulle.
\newblock A fast iterative shrinkage-thresholding algorithm for linear inverse
  problems.
\newblock \emph{{SIAM} Journal on Imaging Sciences}, 2\penalty0 (1):\penalty0
  183--202, 2009.

\bibitem[Brown et~al.(2020)Brown, Mann, Ryder, Subbiah, Kaplan, Dhariwal,
  Neelakantan, Shyam, Sastry, Askell, Agarwal, Herbert-Voss, Krueger, Henighan,
  Child, Ramesh, Ziegler, Wu, Winter, Hesse, Chen, Sigler, Litwin, Gray, Chess,
  Clark, Berner, McCandlish, Radford, Sutskever, and Amodei]{BMR+20}
Tom Brown, Benjamin Mann, Nick Ryder, Melanie Subbiah, Jared Kaplan, Prafulla
  Dhariwal, Arvind Neelakantan, Pranav Shyam, Girish Sastry, Amanda Askell,
  Sandhini Agarwal, Ariel Herbert-Voss, Gretchen Krueger, Tom Henighan, Rewon
  Child, Aditya Ramesh, Daniel Ziegler, Jeffrey Wu, Clemens Winter, Chris
  Hesse, Mark Chen, Eric Sigler, Mateusz Litwin, Scott Gray, Benjamin Chess,
  Jack Clark, Christopher Berner, Sam McCandlish, Alec Radford, Ilya Sutskever,
  and Dario Amodei.
\newblock Language models are few-shot learners.
\newblock In \emph{Advances in Neural Information Processing Systems}, 2020.

\bibitem[Chen et~al.(2019)Chen, Li, Tao, Barnett, Su, and Rudin]{CLT+19}
Chaofan Chen, Oscar Li, Chaofan Tao, Alina~Jade Barnett, Jonathan Su, and
  Cynthia Rudin.
\newblock This looks like that: Deep learning for interpretable image
  recognition.
\newblock In \emph{Neural Information Processing Systems}, 2019.

\bibitem[Cohen and Welling(2016)]{CoWe16}
Taco Cohen and Max Welling.
\newblock Group equivariant convolutional networks.
\newblock In \emph{International Conference on Machine Learning}, 2016.

\bibitem[Devlin et~al.(2019)Devlin, Chang, Lee, and Toutanova]{DCL+19}
Jacob Devlin, Ming-Wei Chang, Kenton Lee, and Kristina Toutanova.
\newblock {BERT:} pre-training of deep bidirectional transformers for language
  understanding.
\newblock In \emph{Conference of the North American Chapter of the Association
  for Computational Linguistics}, 2019.

\bibitem[Dieleman et~al.(2016)Dieleman, De~Fauw, and Kavukcuoglu]{DDK16}
Sander Dieleman, Jeffrey De~Fauw, and Koray Kavukcuoglu.
\newblock Exploiting cyclic symmetry in convolutional neural networks.
\newblock In \emph{International Conference on Machine Learning}, 2016.

\bibitem[Esteves et~al.(2020)Esteves, Makadia, and Daniilidis]{EMD20}
Carlos Esteves, Ameesh Makadia, and Kostas Daniilidis.
\newblock Spin-weighted spherical cnns.
\newblock In \emph{Neural Information Processing Systems}, 2020.

\bibitem[Gregor and LeCun(2010)]{GrLe10}
Karol Gregor and Yann LeCun.
\newblock Learning fast approximations of sparse coding.
\newblock In \emph{International Conference on Machine Learning}, 2010.

\bibitem[He et~al.(2016)He, Zhang, Ren, and Sun]{HZRS16}
Kaiming He, Xiangyu Zhang, Shaoqing Ren, and Jian Sun.
\newblock Deep residual learning for image recognition.
\newblock In \emph{IEEE Conference on Computer Vision and Pattern Recognition},
  2016.

\bibitem[Ioffe and Szegedy(2015)]{IoSz15}
Sergey Ioffe and Christian Szegedy.
\newblock Batch normalization: Accelerating deep network training by reducing
  internal covariate shift.
\newblock In \emph{International Conference on Machine Learning}, 2015.

\bibitem[Marcos et~al.(2017)Marcos, Volpi, Komodakis, and Tuia]{MVKT17}
Diego Marcos, Michele Volpi, Nikos Komodakis, and Devis Tuia.
\newblock Rotation equivariant vector field networks.
\newblock In \emph{International Conference on Computer Vision}, 2017.

\bibitem[Mildenhall et~al.(2020)Mildenhall, Srinivasan, Tancik, Barron,
  Ramamoorthi, and Ng]{MST+20}
Ben Mildenhall, Pratul Srinivasan, Matthew Tancik, Jonathan Barron, Ravi
  Ramamoorthi, and Ren Ng.
\newblock {NeRF:} representing scenes as neural radiance fields for view
  synthesis.
\newblock In \emph{European Conference on Computer Vision}, 2020.

\bibitem[Nguyen et~al.(2019)Nguyen, Wong, and Hegde]{NWH19}
Thanh Nguyen, Raymond Wong, and Chinmay Hegde.
\newblock On the dynamics of gradient descent for autoencoders.
\newblock In \emph{International Conference on Artificial Intelligence and
  Statistics}, 2019.

\bibitem[Pavlakos et~al.(2017)Pavlakos, Zhou, Derpanis, and Daniilidis]{PZDD17}
Georgios Pavlakos, Xiaowei Zhou, Konstantinos Derpanis, and Kostas Daniilidis.
\newblock Coarse-to-fine volumetric prediction for single-image {3D} human
  pose.
\newblock In \emph{IEEE Conference on Computer Vision and Pattern Recognition},
  2017.

\bibitem[Radford et~al.(2021)Radford, Kim, Hallacy, Ramesh, Goh, Agarwal,
  Sastry, Askell, Mishkin, Clark, Krueger, and Sutskever]{RKH+21}
Alec Radford, Jong~Wook Kim, Chris Hallacy, Aditya Ramesh, Gabriel Goh,
  Sandhini Agarwal, Girish Sastry, Amanda Askell, Pamela Mishkin, Jack Clark,
  Gretchen Krueger, and Ilya Sutskever.
\newblock Learning transferable visual models from natural language
  supervision.
\newblock OpenAI, 2021.

\bibitem[Redmon and Farhadi(2017)]{ReFa17}
Joseph Redmon and Ali Farhadi.
\newblock {YOLO9000:} better, faster, stronger.
\newblock In \emph{IEEE Conference on Computer Vision and Pattern Recognition},
  2017.

\bibitem[Ribeiro et~al.(2016)Ribeiro, Singh, and Guestrin]{RSG16}
Marco~Tulio Ribeiro, Sameer Singh, and Carlos Guestrin.
\newblock "why should i trust you?" explaining the predictions of any
  classifier.
\newblock In \emph{International Conference on Knowledge Discovery and Data
  Mining}, 2016.

\bibitem[Shlezinger et~al.(2020)Shlezinger, Whang, Eldar, and Dimakis]{SWED20}
Nir Shlezinger, Jay Whang, Yonina Eldar, and Alexandros Dimakis.
\newblock Model-based deep learning.
\newblock In \emph{arXiv}, 2020.

\bibitem[Simon and Elad(2019)]{SiEl19}
Dror Simon and Michael Elad.
\newblock Rethinking the {CSC} model for natural images.
\newblock In \emph{Neural Information Processing Systems}, 2019.

\bibitem[Sulam et~al.(2020)Sulam, Aberdam, Beck, and Elad]{SABE20}
Jeremias Sulam, Aviad Aberdam, Amir Beck, and Michael Elad.
\newblock On multi-layer basis pursiot, efficient algorithms and convolutional
  neural netwroks.
\newblock \emph{IEEE Transactions on Pattern Analysis and Machine Learning},
  42\penalty0 (8):\penalty0 1968--1980, 2020.

\bibitem[Tolooshams et~al.(2020)Tolooshams, Song, Temereanca, and Ba]{TST+20}
Bahareh Tolooshams, Andrew Song, Simona Temereanca, and Demba Ba.
\newblock Convolutional dictionary learning based auto-encoders for natural
  exponential-family distributions.
\newblock In \emph{International Conference on Machine Learning}, 2020.

\bibitem[Tolooshams et~al.(2021)Tolooshams, Dey, and Ba]{TDB21}
Bahareh Tolooshams, Sourav Dey, and Demba Ba.
\newblock Deep residual autoencoders for expectation maximization-inspired
  dictionary learning.
\newblock \emph{IEEE Transactions on Neural Networks and Learning Systems},
  32\penalty0 (6):\penalty0 2415--2429, 2021.

\bibitem[Worrall et~al.(2017)Worrall, Garbin, Turmukhambetov, and
  Brostow]{WGTB17}
Daniel Worrall, Stephan Garbin, Daniyar Turmukhambetov, and Gabriel Brostow.
\newblock Harmonic networks: Deep translation and rotation equivariance.
\newblock In \emph{IEEE Conference on Computer Vision and Pattern Recognition},
  2017.

\end{thebibliography}
\newpage
\appendix
\section{Learned filters on CIFAR-10}
\label{apd:first}
We present filters learned on CIFAR-10 (without whitening) by the three architectures. We choose to present filters from the first layer, as those resemble edge detectors the most and thus are more straightforward to reason about.
\begin{figure}[h!]
    \centering
    \includegraphics[scale=0.5]{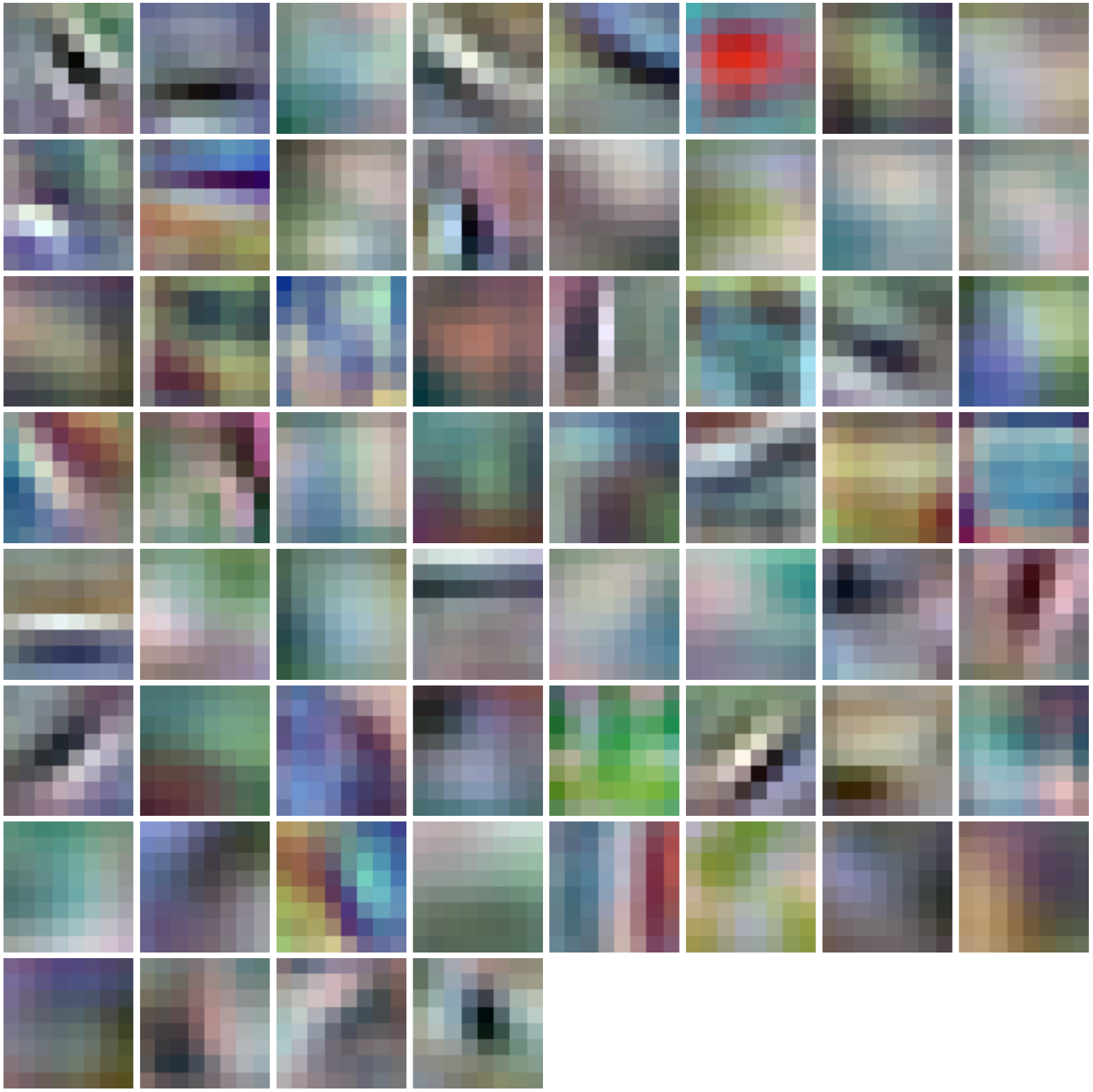}
    \caption{Filters learned using the baseline architecture.}
    \label{fig:apnd1}
\end{figure}
\begin{figure}[h!]
    \centering
    \includegraphics[scale=0.5]{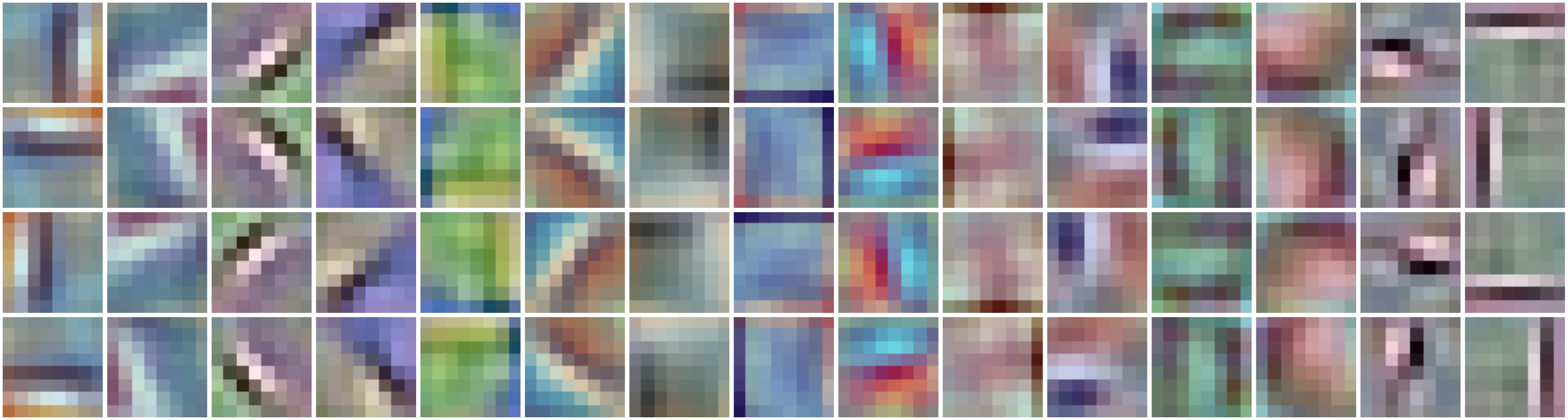}
    \caption{Filters learned using $R_{90}$-CNN.}
    \label{fig:apnd2}
\end{figure}
\begin{figure}[h!]
    \centering
    \includegraphics[scale=0.5]{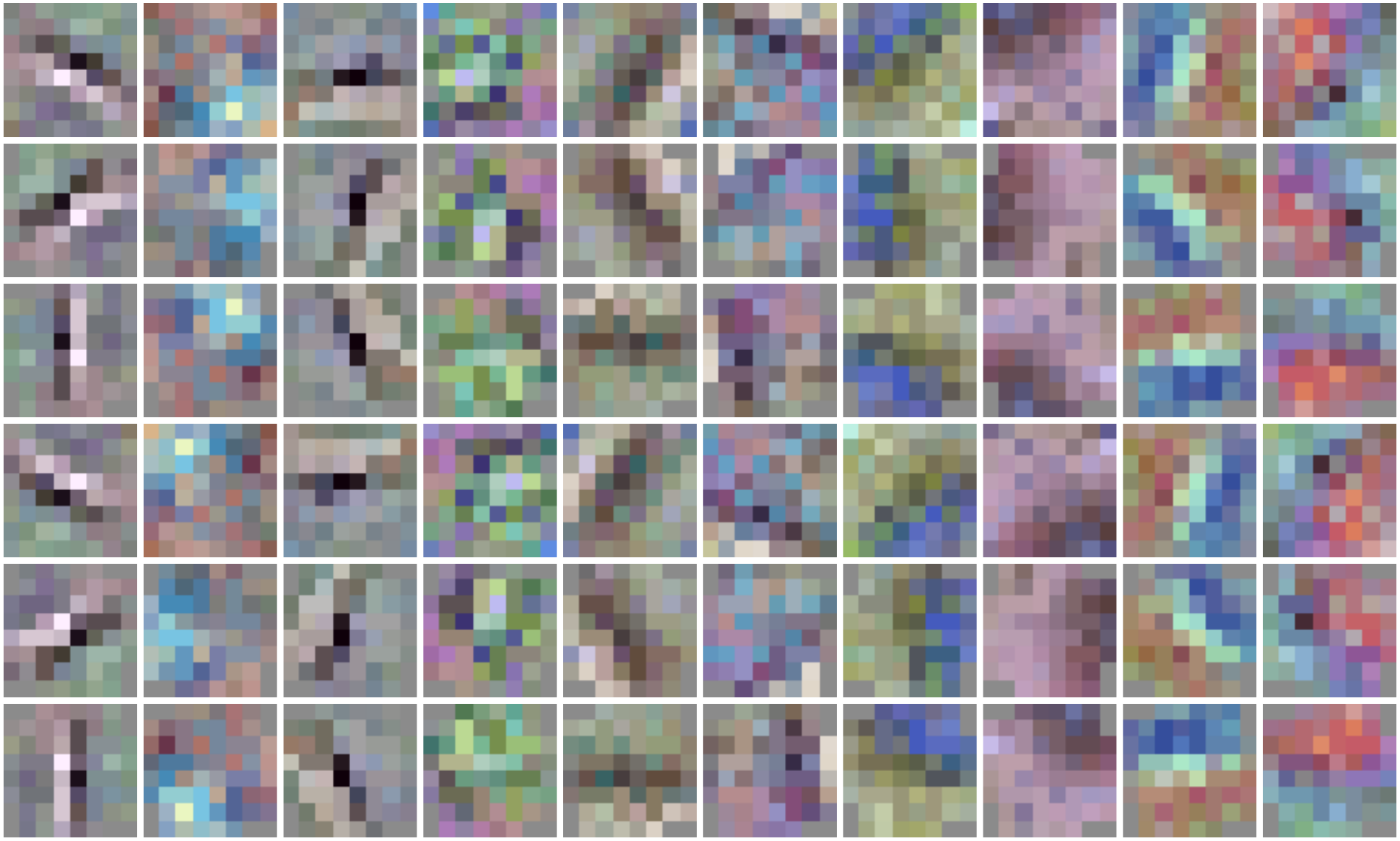}
    \caption{Filters learned using $R_{60}$-CNN.}
    \label{fig:apnd3}
\end{figure}
While all models seem to be learning similarly-looking filters, the equivariant models do not need to ``waste'' computation on learning different orientations. Indeed, if we observe \cref{fig:apnd1}, the first filter from the top and the third from the bottom of the first column seem to be rotated versions of one another. In stark contrast, the third column of \cref{fig:apnd2} seems to learning the elementary rotations of that same filter, without investing resources on learning that information from the data. That is also observed in the first column of \cref{fig:apnd3}, which has even more rotated versions of that same filter.
\end{document}